\documentclass{article}

\PassOptionsToPackage{numbers, compress}{natbib}

\usepackage[final]{nips_2018}

\usepackage[utf8]{inputenc} 
\usepackage[T1]{fontenc}    
\usepackage{hyperref}       
\usepackage{url}            
\usepackage{booktabs}       
\usepackage{amsfonts}       
\usepackage{nicefrac}       
\usepackage{microtype}      
\usepackage{amsmath}
\usepackage{epsf}
\usepackage[hang]{subfigure}
\usepackage{graphicx}

\DeclareMathOperator*{\argmax}{arg\,max}
\DeclareMathOperator*{\argmin}{arg\,min}

\usepackage{floatrow}
\newfloatcommand{capbtabbox}{table}[][\FBwidth]

\title{Risk-averse Behavior Planning for Autonomous Driving under Uncertainty}

\author{
  Mohammad Naghshvar \quad Ahmed K. Sadek \quad Auke J. Wiggers\\
  Qualcomm AI Research and Qualcomm Automotive Research\\
   \texttt{\{mnaghshvar, asadek, auke\}@qualcomm.qti.com} \\
}

\begin{document}

\maketitle

\begin{abstract}
Autonomous vehicles have to navigate the surrounding environment with partial observability of other objects sharing the road. Sources of uncertainty in autonomous vehicle measurements include sensor fusion errors, limited sensor range due to weather or object detection latency, occlusion, and hidden parameters such as other human driver intentions. Behavior planning must consider all sources of uncertainty in deciding future vehicle maneuvers. This paper presents a scalable framework for risk-averse behavior planning under uncertainty by incorporating QMDP, unscented transform, and Monte Carlo tree search (MCTS). It is shown that upper confidence bound (UCB) for expanding the tree results in noisy Q-value estimates by the MCTS and a degraded performance of QMDP. A modification to action selection procedure in MCTS is proposed to achieve robust performance.
\end{abstract}

\section{Introduction}

A major challenge for autonomous driving is achieving safe and comfortable driving experience under uncertainty. Several factors contribute to the uncertainty in the road world model (RWM) around the ego-vehicle (EV), which can be divided into uncertainty due to measurement noise and uncertainty associated with the environment. Examples for measurement noise include inherent sensor noise, errors in perception pipelines (e.g., object detection, road segmentation), limited sensor range due to weather or latency in detecting/tracking stationary/slowly-moving objects or objects that suddenly appear from occluded areas. Examples for uncertainty from the environment are hidden RWM states due to occlusions from large obstacles or sharp turns, or parameters that cannot be physically sensed such as intentions of humans participating in the environment, including other drivers, cyclists, and pedestrians.

Different approaches have been proposed to solve the behavior planning problem. Handwritten rules using finite state machines have been used in the DARPA Urban drive competition \cite{DARPA}, e.g., \cite{Urmson2008}. Another approach is to  formulate the problem as a partially observable Markov decision process (POMDP) where the uncertainty in the environment is captured by a belief about its true state. POMDP is a powerful framework for solving the behavior planning, and several papers have been published using approximate POMDP solvers, e.g., \cite{Bandyopadhyay2013, Brechtel2011, Brechtel2014, Cunningham2015, Ulbrich2013}.

This paper presents risk-averse QMDP (RA-QMDP) --- a scalable approach for decision making under uncertainty by incorporating QMDP \cite{QMDP}, unscented transform \cite{UT}, and MCTS \cite{MykelBook}. QMDP is a method for solving a POMDP by sampling from the belief and leveraging the solution to the underlying MDP; while unscented transform (also referred to as sigma-point sampling) selects samples and their corresponding weights from a given distribution in a way that the original mean/covariance is preserved. Together they provide an efficient tool to sample from the belief, evaluate mean/variance of the outcome for different maneuvers, and make a risk-averse decision. To solve the underlying MDP, we propose a modified MCTS approach which performs a comprehensive evaluation of all permissible maneuvers (such as lane keeping or lane change negotiation/execution with various configurable parameters) over a sufficiently long planning horizon.

The approach presented in this paper can be easily extended to different driving scenarios, and in this paper we show results for two important use cases for highway driving. First use case is behavior planning under limited sensor range where we show that the proposed RA-QMDP approach achieves a tradeoff between EV's average velocity and worst-case absolute jerk as a function of sensor range. The second use case is on handling a merging car that is partially observed due to occlusion. In both cases, we show that the proposed algorithm is able to achieve a tradeoff between safety and assertiveness.

\section{Related Work}

A tree search based approach for behavior planning has been proposed in \cite{Zoox} where a deep reinforcement learning (DRL) agent is trained to optimize the tree search efficiency. The paper however does not address uncertainty and assumes a fully observed environment (MDP).
\cite{Litkouhi2011} proposes to use QMDP for single lane autonomous driving under uncertainty. The paper assumes a Gaussian model for sensor noise and uses normalized probability density values as weights. Since actual noise is not Gaussian and the transformation from observation to decision is non-linear, this model assumption may be prone to mismatch errors. \cite{Lenz2016} frames highway planning as a cooperative perfect information game, and proposes to use MCTS to minimize a global, shared cost function.
\cite{Litkouhi2017} addresses the ramp merging scenario using a probabilistic graphical model.

\cite{Sunberg2017} proposes POMCPOW and PFT-DPW as extensions of MCTS to POMDP settings with continuous observation and action spaces. In particular, the paper applies this problem to lane changing, and uses a particle filter to track predictions for parameters used to model driver intentions (assuming remaining observations are ideal).
\cite{IS-DSPOT} considers importance sampling with a tree search algorithm, where samples are obtained using a probability distribution.
\cite{Sunberg2018_BRL} considers the joint estimation and control problem for a robot using an unscented Kalman filter \cite{UKF} to estimate different parameter values, and a variation of QMDP tree search that assumes full observability of the environment after the first step for computation efficiency.
\cite{BMW} proposes a POMDP planning framework to handle automated driving at an unsignalized intersection where a particle filter is used to represent belief about the routes other vehicles may take. They sample from the particles and evaluate various longitudinal acceleration along EV's path and select the action achieving the highest expected reward at the end.

As we will explain in more detail in the next sections, our main contributions to the prior work discussed here is introducing a sample-efficient framework that enables risk-averse decision making by tracking the uncertainty (in addition to the mean) of the total reward that an action can achieve. Moreover, rare critical events can be examined more thoroughly in this framework as they will have their corresponding sigma-point and a dedicated tree search.

\section{Approach}

A modular stack is considered where a RWM is inferred based on sensor inputs and map information. Let $\mathbf{s}_t$ represent the state of the EV as well as other objects sharing the road with EV at time $t$, where $j$-th object state is defined as
\begin{equation}
\mathbf{s}_t^j=\left[x^j, y^j, v_x^j, v_y^j, a_x^j, a_y^j \right], 
\end{equation}
which represent lateral and longitudinal location, velocity, and acceleration 
 for the object, respectively, and time indices are dropped for simplicity. The RWM module generates a mean $\mu_\mathbf{s}$ and covariance matrix $\Sigma_\mathbf{s}$ that are calculated by the internal tracking algorithm.

A hierarchical behavior planner is considered where the action space, $\mathcal{A}$, is defined as a composition of a high level action set $A_{HL}=\left\{\text{LaneKeep}, \text{LaneChange\textsubscript{R}}, \text{LaneChange\textsubscript{L}}, \text{Yield}, \text{Negotiate\textsubscript{lanechange}}\right\}$ and a set of parameters, $\Theta$, specifying how to execute the action (e.g., $\Theta$ may include safe time headway, minimum distance from a lead vehicle, desired velocity, maximum acceleration/deceleration, level of politeness, and direction and maximum time/distance of lane change). 

The behavior planner action is then passed to a motion planning and control module to generate the trajectory planning and throttle/steering control. Based on a time series of inputs from the RWM, the behavior planner selects an action based on a risk-averse metric:
\begin{equation}
\label{eqn-risk-averse}
a^*=\argmax_{a \in \mathcal{A}} \left(Q\left(b\left(\mathbf{s}\right),a\right)-\alpha \times \sigma^2\left(b\left(\mathbf{s}\right),a\right)\right),
\end{equation}
where $b\left(\mathbf{s}\right)$ is the belief about true state $\mathbf{s}$ which is represented by a mean $\mu_\mathbf{s}$ and covariance function $\Sigma_\mathbf{s}$, $Q\left(b\left(\mathbf{s}\right),a\right)$ is the expected value of taking action $a$ in belief state $b\left(\mathbf{s}\right)$, $\sigma^2\left(b\left(\mathbf{s}\right), a\right)$ represents the variance of this value, and $\alpha$ is a hyperparameter.
This is a POMDP problem and its exact solution is computationally intractable. Instead, we use a QMDP-based framework equipped with the unscented transform to generate sigma-points to sample the belief distribution. In particular, we first select $2n_\mathbf{s}+1$ sigma-points in a deterministic way as follows:
\begin{equation}
\label{eqn-ut}
\mathbf{x}_0=\mu_\mathbf{s}, \quad \mathbf{x}_{i}=\mu_\mathbf{s}+ \left(\sqrt{\frac{n_\mathbf{s}}{1-W_0}\Sigma_\mathbf{s}}\right)_i, \quad  \mathbf{x}_{i+n_\mathbf{s}}=\mu_\mathbf{s}- \left(\sqrt{\frac{n_\mathbf{s}}{1-W_0} \Sigma_\mathbf{s}}\right)_i,
\end{equation}
where $n_\mathbf{s}$ is the dimension of the vector $\mathbf{s}$, $i$ takes values in $[1, 2, \ldots, n_\mathbf{s}]$, $W_0$ is the weight corresponding to $\mathbf{x}_0$ that can be positive or negative, and $\left(\sqrt{\frac{n_\mathbf{s}}{1-W_0}\Sigma_\mathbf{s}}\right)_i$ is the $i$-th row or column of the matrix square root of $\frac{n_\mathbf{s}}{1-W_0}\Sigma_\mathbf{s}$. Weights corresponding to other sigma-points are the same and are given by $W_i=\left(1-W_0\right)/2n_\mathbf{s}$. In our implementation, any sigma-point that is too close to $\mathbf{x}_0$ or results in physically infeasible situation (e.g., a car is placed out of the road, overlaps with another object, or has acceleration/deceleration/velocity beyond physical limits) is excluded. Therefore, the final value of $n_\mathbf{s}$ used in equation~(\ref{eqn-ut}) could be smaller than the dimension of vector $\mathbf{s}$.
The mean and variance of the Q-function are then calculated as follows:
\begin{eqnarray}
\label{eqn-qmdp-mean}
  \hat{Q}\left(b\left(\mathbf{s}\right),a\right) &=& \sum_{i=1}^{2n_\mathbf{s}+1}W_i Q_{MDP}\left(\mathbf{x}_i, a\right),\\
   \label{eqn-qmdp-var}
  \hat{\sigma}^2\left(b\left(\mathbf{s}\right),a\right) &=& \sum_{i=1}^{2n_\mathbf{s}+1}W_i \left( Q_{MDP}\left(\mathbf{x}_i, a\right)-\hat{Q}\left(b\left(\mathbf{s}\right),a\right) \right)^2.
\end{eqnarray}

To calculate $Q_{MDP}\left(\mathbf{s}, a\right)$ for a given state $\mathbf{s}$ we use an online planner, in particular a Monte Carlo tree search (MCTS). In baseline MCTS, the tree is expanded using the UCT bound \cite{MykelBook} as follows:
\begin{equation}
a^*=\argmax_{a\in\mathcal{A}} \left(Q\left(\mathbf{s},a\right) + C\sqrt{\frac{\log\left(N(\mathbf{s})\right)}{N(\mathbf{s},a)}}\right),
\end{equation}
where $N(\mathbf{s},a)$ is the number of times action $a$ was selected when state $\mathbf{s}$ was visited, $N(\mathbf{s})$ is total number of visits to state $\mathbf{s}$, and $C$ is a hyperparameter that tunes the exploration/exploitation tradeoff. It should be noted that the main objective of the MCTS approach is to select the best action $a^*$, and hence ranking of the Q-values across different actions is sufficient, and accuracy of the value function is not very important, especially for actions with low values.

This is troublesome for the QMDP estimates given by equations~(\ref{eqn-qmdp-mean}) and~(\ref{eqn-qmdp-var}), since an action that leads to a low value for a given sigma-point, can lead to a high value for a different sigma-point. We found that using the UCT bound can result in very few visits for actions with low values, resulting in unreliable estimate for the corresponding Q-value. To mitigate this problem, we incorporated stronger exploration approach into MCTS, by combining the UCT bound with $\epsilon$-greedy exploration. It is also possible to add a variance term to the UCT bound for more efficient exploration. In this paper, we implemented the $\epsilon$-greedy approach since it resulted in acceptable performance. The modified action selection criteria is then given by
\begin{equation}
  a^* =
    \begin{cases}
      \argmax_{a\in\mathcal{A}} \left(Q\left(\mathbf{s},a\right) + C\sqrt{\frac{\log\left(N(\mathbf{s})\right)}{N(\mathbf{s},a)}}\right), & \text{with probability $1-\epsilon$}\\
      \argmin_{a\in\mathcal{A}} N(\mathbf{s},a), & \text{otherwise}\\
    \end{cases}.       
\end{equation}
For efficient tree search, we only limit using the $\epsilon$-greedy for selection of actions at the root of the tree, and use UCT for selecting actions at other nodes.

Finally, we should point out that our method also extends to cases where the belief state contains both continuous and discrete random variables. For example continuous noisy values for lateral/longitudinal location/velocity/acceleration, and discrete/categorical values for blinker status, type (car, truck, van, etc.), and absence/presence of an object at sensor range. For the discrete or categorical random variables we use all the non-zero probability realizations along with the potential sigma-points selected for the continuous ones. 

\section{Simulation Results}
In this section, we investigate two driving scenarios with uncertainty: 1) stationary object on the road beyond sensor range, and 2) highway ramp with limited field of view. Before we proceed, we provide a few details which are common to both scenarios. It is assumed that all vehicles use an intelligent driver model (IDM) \cite{IDM} to set their longitudinal acceleration:
\begin{equation}
a = a_{\text{max}} \left[1 - \left(\frac{v}{v_{\text{desired}}}\right)^4 - \left(\frac{s^*(v, v_{\text{lead}})}{s}\right)^2  \right],
\end{equation}
where $v$ and $s$ are velocity of the vehicle and the distance to its lead, respectively, and the safe distance $s^*$ is set according to Lemma~2 in \cite{Mobileye}. Table~\ref{table-params} provides further details about the model parameters.
\begin{equation}
s^*(v, v_{\text{lead}}) = \max \left\{ s_0, v\rho + \frac{1}{2} a_{\text{max}} \rho^2 + \frac{(v+\rho a_{\text{max}})^2}{2 b_{\text{safe}}} - \frac{v_{\text{lead}}^2}{2 b_{\text{max}}} \right\}.
\end{equation}

\begin{table}
  \caption{Model parameters used throughout this paper.}
  \label{table-params}
  \centering
  \begin{tabular}{lll}
    \toprule
    Name                 & Description                        & Value \\
    \midrule
    $s_0$                & Minimum distance in jammed traffic & $2$m     \\
    $\rho$               & Response time                      & $0.25$s  \\
    $v_{\text{desired}}$ & Desired velocity                   & $105$km/h ($29.17$m/s) \\
    $a_{\text{max}}$     & Maximum acceleration               & $2$m/s\textsuperscript{2} \\
    $b_{\text{safe}}$    & Safe deceleration                  & $4$m/s\textsuperscript{2} \\
    $b_{\text{max}}$     & Maximum deceleration               & $8$m/s\textsuperscript{2} \\
    \bottomrule
  \end{tabular}
\end{table}

The behavioral planning (BP) module operates at 2Hz. All algorithms studied in the next subsections perform a tree search of depth 15 (which implies 7.5s planning horizon). The number of tree queries is limited to 20K which is uniformly distributed over sample(s) from the (belief) state.
The action space consists of the LaneKeep action with five configurations that put different lower and upper bounds on the longitudinal acceleration. In particular, we have the following intervals for the acceleration: [-8, -2], [-2, -1], [-1, 0], [0, 1], [1, 2]. For rollout policy during tree search we use an IDM with acceleration restricted to the interval [-8, 0]. This rollout policy allows the EV to continue at constant velocity ($\leq v_\text{desired}$) and keep a safe distance to its lead vehicle; hence it returns a reasonable estimate for the Q-value update. 
State transitions during the tree search are modeled via a basic prediction algorithm that has perfect knowledge about EV's dynamics while assumes other vehicles drive at constant velocity. 
The cost function is a weighted sum of three components which focus on safety, comfort, and performance.  For safety, we consider penalty for closeness to other vehicles and crash (as a function of the velocity of the involved vehicles); for comfort, separate cost functions for hard brake and jerk are defined; and finally for performance, deviation from desired velocity is penalized.
After the tree search finishes, the best action is selected according to equation (\ref{eqn-risk-averse}) and is sent to the motion planning (MP) module that is running at 20Hz. The MP module respects the upper bound specified by the BP action but it may override the lower bound in case it finds higher deceleration necessary to ensure safety.

\subsection{Stationary Object on the Road}
\begin{figure}
\begin{floatrow}
\ffigbox{%
  \centering
     \includegraphics[width=6.5cm, trim={0 3.5cm 0 0},clip]{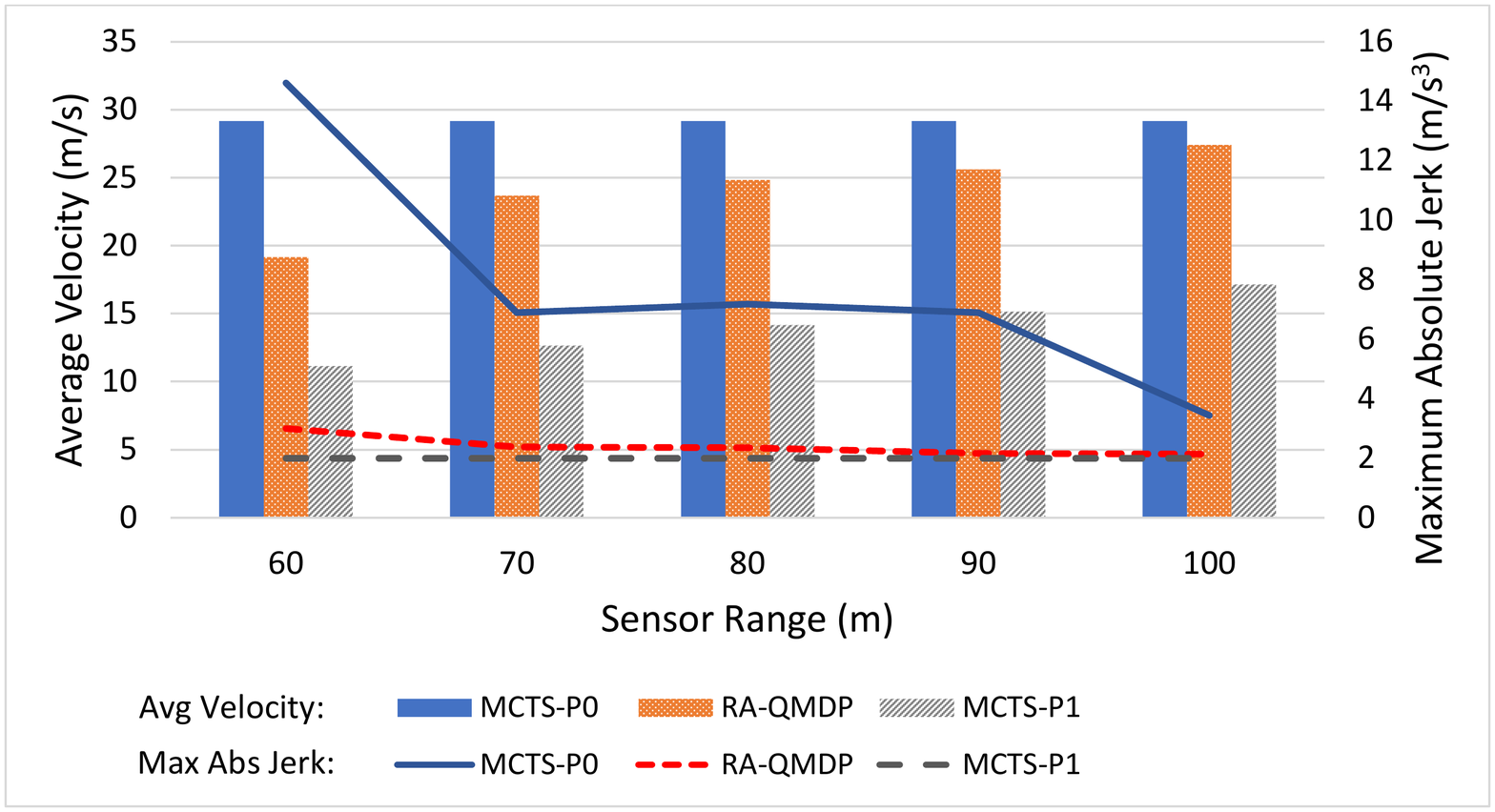}
}{%
  \caption{Velocity-Jerk tradeoff as sensor range varies.}%
  \label{fig-exp1-comparison-to-baseline}
}
\capbtabbox{%
  \begin{tabular}{ccccc} 
  \toprule
  $\alpha$ & $\epsilon$ & $\bar{v}_{EV}$ & $s^*(\bar{v}_{EV},0)$ & abs jerk \\
  \midrule
  $0.1$ & $1.0$ & $18.93$ & $51.99$ & $4$ \\
  $0.01$ & $1.0$ & $19.17$ & $53.19$ & $3$ \\
  $0.0$ & $1.0$ & $23.17$ & $75.86$ & $3.6$ \\
  \midrule
  $0.01$ & $0.5$ & $13.67$ & $28.58$ & $2.7$ \\ 
  $0.01$ & $0.2$ & $16.42$ & $39.95$ & $4.1$ \\
  $0.01$ & $0.0$ & $16.48$ & $40.22$ & $5.5$ \\
  \bottomrule
  \end{tabular}
}{%
  \caption{Performance of {RA-QMDP}{$(\alpha$,$\epsilon)$} for sensor range $60$m as $\alpha$ and $\epsilon$ vary.}%
  \label{table-exp1-riskaverse-benefit}
}
\end{floatrow}
\end{figure}

The experiment presented in this section addresses the sensor range limitation in autonomous driving which could be due to poor weather/light condition, occlusion by large obstacles, difficulty in detecting stationary objects on the road, etc.  
In this experiment, we assume there exists a stationary object on EV's path, $400$m away from its current location. Let us define sensor range as a distance at which the probability of detecting a stationary object is $10\%$. In this scenario, our proposed RA-QMDP($\alpha,\epsilon$) approach with risk-averse coefficient $\alpha=0.01$ and $\epsilon$-greedy parameter $\epsilon=1$ takes two samples: 1)~a stationary object is present at the sensor range, 2) the road ahead is clear of any object; where the corresponding weights for these two samples are $0.1$ and $0.9$, respectively.
For baseline we consider two MCTS-based schemes, denoted by MCTS-P0 and MCTS-P1, 
that assume there is a stationary object on the road at sensor range with probability $0$ and $1$, respectively. 
Figure~\ref{fig-exp1-comparison-to-baseline} compares the average velocity and the maximum absolute jerk experienced by the candidate algorithms for this experiment. MCTS-P0 is aggressive and achieves the highest average velocity (equal to $v_{\text{desired}}$). However, when the stationary object appears in the sensor range, this algorithm experiences the worst absolute jerk as it needs to decelerate immediately to avoid an accident.\footnote{With the parameters given in Table~\ref{table-params}, MCTS-P0 cannot avoid crash for a sensor range less than $53$m.} MCTS-P1 on the other hand is conservative and settles at a much lower velocity. As a result, this algorithm experiences the lowest absolute jerk because it has sufficient time in order to decelerate and stop behind the stationary object. RA-QMDP is able to achieve a tradeoff between safety and performance. 

To illustrate the advantage of selecting a risk-averse decision as well as comprehensive exploration at the tree root, we compare variants of {RA-QMDP}{$(\alpha$,$\epsilon)$} with risk-averse coefficient $\alpha \in \{0, 0.01, 0.1\}$ and $\epsilon$-greedy parameter $\epsilon \in \{0, 0.2, 0.5, 1.0\}$.
Table~\ref{table-exp1-riskaverse-benefit} shows the average velocity, its corresponding safe distance to a stationary object, and worst case absolute jerk achieved by each of these variants of RA-QMDP for a scenario where sensor range is $60$m. Choosing $\alpha=0$ implies selecting the action with highest Q-value which is the most common procedure that MCTS- and QMDP-based solvers apply.
As shown in Table~\ref{table-exp1-riskaverse-benefit}, this selection results in a risky performance where the safe distance is larger than the sensor range. Selecting a very large value for ${\alpha}$ is also not ideal and impacts robustness as variance term dominates the mean in (\ref{eqn-risk-averse}). 
Selecting $\epsilon$ close to 0 results in insufficient explorations of available actions at the root of the tree and hence, an unstable performance in various sensor ranges. For the specific example shown in Table~\ref{table-exp1-riskaverse-benefit}, and with choice of $\epsilon\in\{0.0, 0.2\}$, EV constantly accelerates and decelerates and hence it could experience a large jerk if the stationary object appears in the sensor range during the acceleration phase.
Selecting $\epsilon$ close to $1$ shows reliable performance in this scenario. In general, however, selecting $\epsilon$ close to 1 implies choosing actions almost at random at the tree root which could be computationally inefficient for high dimensional problems when action space is large. 
Using an adaptive exploration strategy, such as one that incorporates a computed variance in order to improve sample efficiency, is a subject for future work.


\subsection{Highway Ramp with Limited Field of View}
\begin{figure}
	\centering
     \includegraphics[height=4cm, trim={0 7.8cm 0 0},clip]{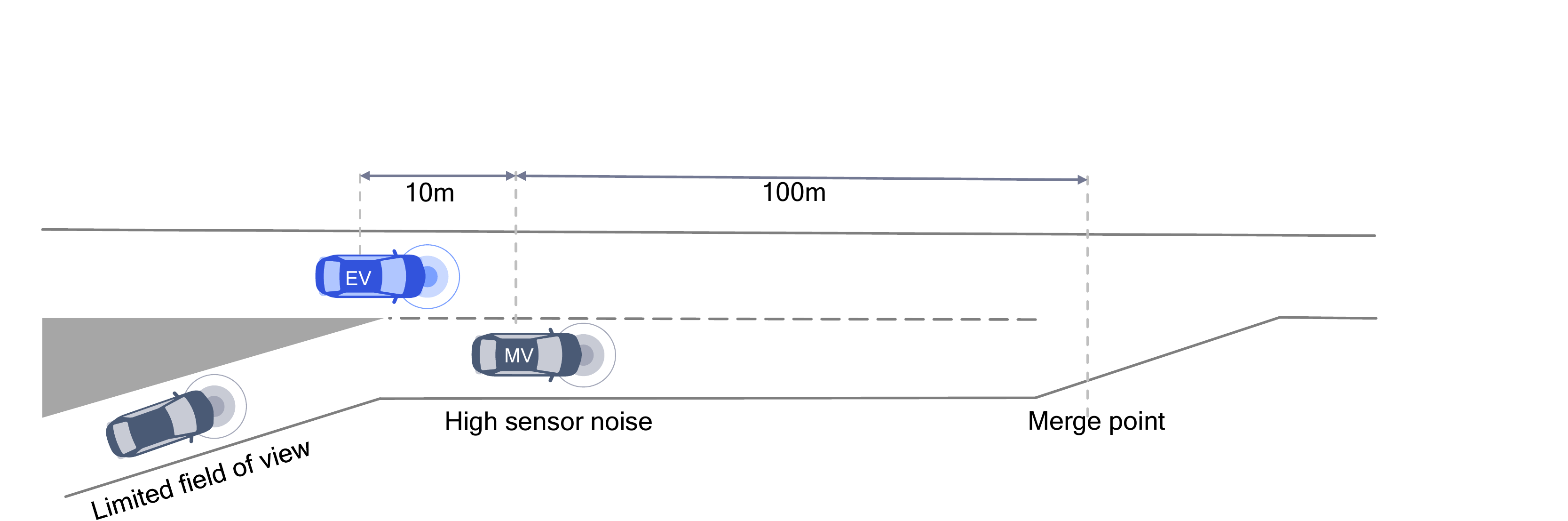}
	\caption{Highway ramp with limited field of view.}
	\label{fig-exp2}
\end{figure}

The second experiment addresses the problem of noisy sensor measurements. The amount of sensor noise could depend on the weather and light conditions, relative distance and orientation with respect to EV, length of time interval the object has been tracked, etc. 
Consider a highway ramp as depicted in Figure~\ref{fig-exp2} where EV experiences a limited field of view due to existing obstacles (e.g., trees) or difference in the elevation.  
As a result, EV's estimate of the state of a merging vehicle (MV) is subject to sensor noise whose value is high initially and decreases gradually as MV is tracked for a longer time. For simplicity of analysis, we assume MV would not switch its lane before the merge point\footnote{The experiment could be easily extended to the case where the behavior prediction module assigns probability to MV's intention of keeping or switching its lane. In that case, there will be multiple sets of sigma-points (similar to the ones derived in this section), one for each probable intention of the MV and the original sigma-point weights are adjusted using the corresponding probability of each intention category.} and there exists noise only in measurements of MV's longitudinal velocity, and all other state elements are measured accurately. 
Let us denote the noise variance reported by RWM at time $t$ by $\sigma^2_N(t)$. To perform unscented transform, we set $W_0=0.5$. According to (\ref{eqn-ut}) we will have three sigma-points at each decision making time $t$ where $\mathbf{x}_0$, $\mathbf{x}_1$, and $\mathbf{x}_2$ are constructed by adding 0, $+\sqrt{2}\sigma_N(t)$, and $-\sqrt{2}\sigma_N(t)$, respectively, to the longitudinal velocity estimates of MV. The corresponding weight for $\mathbf{x}_1$ and $\mathbf{x}_2$ is 0.25.   


\begin{figure}
\begin{floatrow}
\ffigbox{%
  \centering
     \includegraphics[width=6.5cm, trim={0 4cm 0 0},clip]{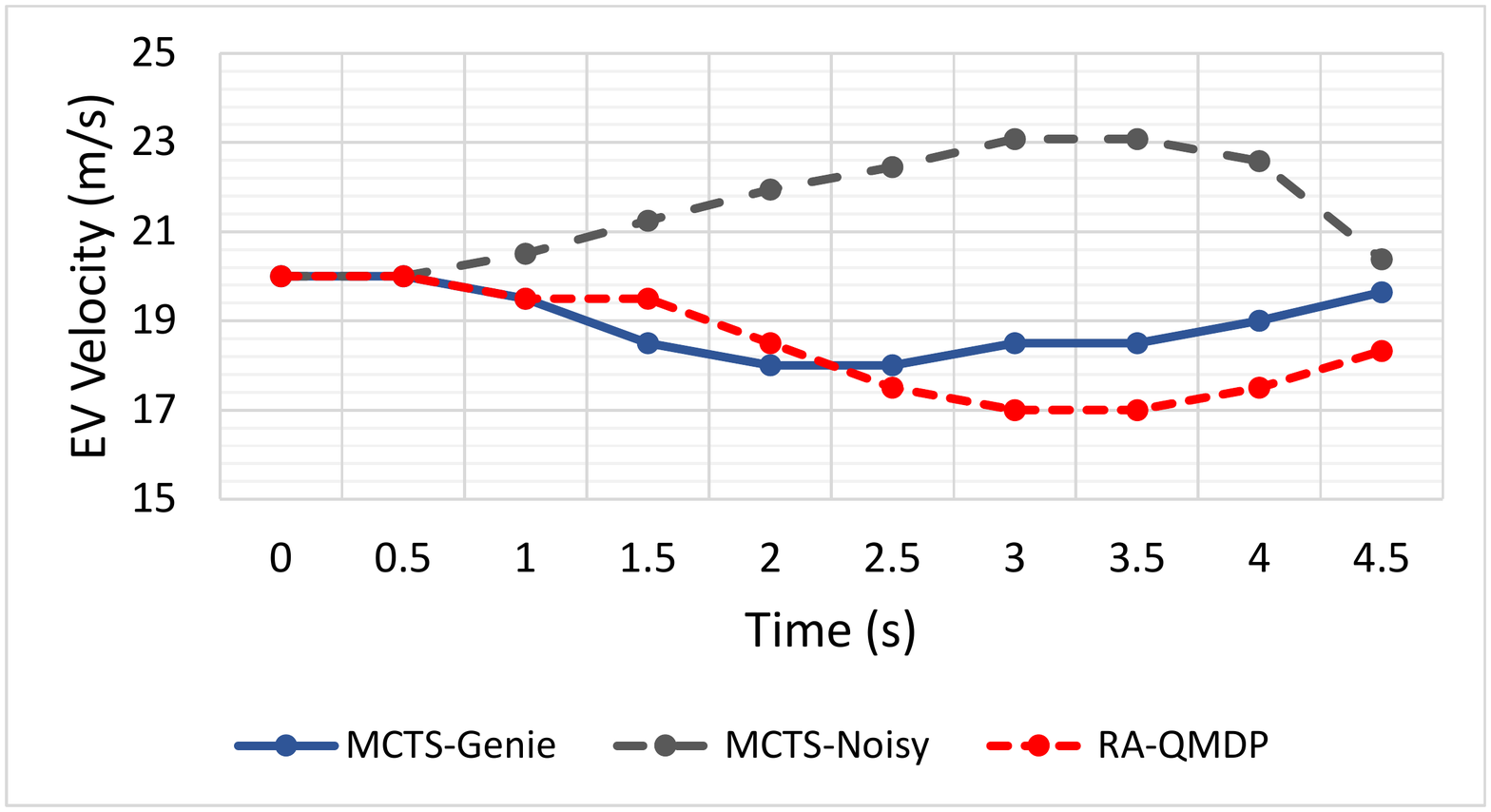}
}{%
  \caption{EV velocity for a realization of noise.}%
  \label{fig-exp2-velocity}
}
\capbtabbox{%
  \begin{tabular}{llll} 
  \toprule
   & \small{MCTS-} & \small{MCTS-} & \small{RA-} \\
   & \small{ Genie} & \small{ Noisy} & \small{QMDP} \\
  \midrule
  \small{Distance to MV} & $28.57$ & $15.8$  & $30.67$ \\
  \small{Time headway}     & $1.27$	 & $0.54$  & $1.41$  \\
  \small{Max abs jerk}     & $2.0$     & $6.8$   & $4.0$ \\
  \small{EV velocity}      & $19.64$ & $20.38$ & $18.32$ \\ 
  \small{MV velocity}      & $25.46$ & $25.46$ & $25.46$ \\
  \bottomrule
  \end{tabular}
}{%
  \caption{Comparison at merge point.}%
  \label{table-exp2-merge-point}
}
\end{floatrow}
\end{figure}

For baseline we consider two MCTS-based schemes: 1) MCTS-Genie that receives noise-free observations, and 2) MCTS-Noisy which makes decisions based on noisy measurements.
The initial velocity of both EV and MV is $20$m/s. 
Figure~\ref{fig-exp2-velocity} compares EV's velocity for the candidate algorithms, and Table~\ref{table-exp2-merge-point} provides further information about the vehicles' status at the merge point
for a realization of the noise under which initial measurements of MV velocity are lower than its actual value.
In the beginning, MCTS-Noisy assumes that MV is going slower than EV, and decides to accelerate to pass MV before reaching the merge point. However, after few seconds, it realizes that the initial measurements were off and in fact it is not able to take over MV. Although it applies aggressive deceleration (with jerk -$6.8$m/s\textsuperscript{3}) at that moment, there is not enough time left to create a safe gap before the merge point (see distance to lead and time headway reported in Table~\ref{table-exp2-merge-point}). MCTS-Genie observes the actual velocity of MV and hence is able to make a correct decision from the beginning. For RA-QMDP, we select $\alpha=0.01$ and $\epsilon=1$. As explained above, RA-QMDP takes three sigma-points one of which considers that MV is going faster than the value that RWM reports. It then makes a risk-averse decision which forces the EV to slow down and increase the gap with MV. Note that RA-QMDP has some delay in making that decision since its evaluation is also influenced by two other sigma-points that suggest MV is going slow and there might be an opportunity to take over MV.   

\section{Conclusion and Future Work}
In this paper, we proposed a sample-efficient framework for behavior planning under uncertainty which incorporates QMDP, unscented transform, and modified MCTS. It was shown how input uncertainty could be propagated to estimate uncertainty in the outcome of candidate maneuvers, hence, enabling risk-averse decision making.

Future area of research includes 1) testing this algorithm on an actual self-driving car, 2) using deep reinforcement learning to improve the efficiency of tree search by pruning the action space, and replacing the rollout procedure, 3) applying an enhanced prediction module with capability of assigning probability to various maneuvers for forward simulations in the tree.


\medskip

\small

\bibliographystyle{abbrvnat}
\bibliography{nips_references}

\end{document}